# Interoceptive machine framework: Toward interoception-inspired regulatory architectures in artificial intelligence

Diego Candia-Rivera

*Sorbonne Université, Paris Brain Institute (ICM), CNRS, INSERM, AP-HP, Hôpital de la Pitié-Salpêtrière, Paris 75013, France*

*Correspondence: diego.candia.r@ug.uchile.cl*

## Abstract

This review proposes an integrative framework grounded on interoception and embodied AI—termed the interoceptive machine framework—that translates biologically inspired principles of internal-state regulation into computational architectures for adaptive autonomy. Interoception, conceived as the monitoring, integration, and regulation of internal signals, has proven relevant for understanding adaptive behavior in biological systems. The proposed framework organizes interoceptive contributions into three functional principles: homeostatic, allostatic, and enactive, each associated with distinct computational roles: internal viability regulation, anticipatory uncertainty-based re-evaluation, and active data generation through interaction. These principles are not intended as direct neurophysiological mappings, but as abstractions that inform the design of artificial agents with improved self-regulation and context-sensitive behavior. By embedding internal state variables and regulatory loops within these principles, AI systems can achieve more robust decision-making, calibrated uncertainty handling, and adaptive interaction strategies, particularly in uncertain and dynamic environments. This approach provides a concrete and testable pathway toward agents capable of functionally grounded self-regulation, with direct implications for human-computer interaction and assistive technologies. Ultimately, the interoceptive machine framework offers a unifying perspective on how internal-state regulation can enhance autonomy, adaptivity, and robustness in embodied AI systems.

# 1. Introduction

In human-machine interaction, endowing machines with human-like behaviors is essential to foster natural, intuitive, and engaging exchanges [1]. Current systems often remain rigid, overly task-focused, and limited in their capacity to adapt to the subtleties of human communication, such as emotions, social cues, and contextual understanding [2]. These shortcomings are particularly evident in sensitive domains like psychological treatments [3], robotic interventions for autism [4], or assistive technologies for vulnerable populations such as dementia [5], where the effectiveness of the interaction depends not only on functionality but also on typically human-associated features such as empathy or meaningful behavior.

This review draws on enactive and autonomy-oriented perspectives from the life sciences, which emphasize that biological agents sustain themselves through ongoing internal regulation and situated interaction with their environments [6]. Concepts such as meaning, adaptivity, and autonomy are used here not as philosophical claims, but as functional descriptors of how living systems integrate internal sensing, prediction, and action [7]. In engineering terms, these perspectives translate into concrete design questions: which internal variables must be monitored, how functional viability is maintained, and how internal state representations should modulate behavior under uncertainty. At present, artificial agents typically lack such self-regulatory capacities, relying instead on externally specified objectives. Biological agents, in contrast, flexibly adapt goals by integrating perceptual, memory-related, and internal-state signals—a capability that highlights the need for AI systems to incorporate analogous forms of internal regulation [8,9]. Although not intended as engineering blueprints, neuroscientific models of consciousness highlight several organizational motifs: a centralized thalamic architecture, a global workspace or an ascending arousal system, that help explain why biological systems integrate internal state monitoring with flexible control [10–12]. Current AI systems lack these motifs [13], which motivates the search for alternative architectures capable of functionally similar regulation.

Embodied AI research shows that robust, adaptive behavior emerges from agents whose cognition is shaped by sensorimotor engagement, bodily constraints, and developmental history [14–19]. These perspectives caution against over-interpreting intelligent behavior while lacking genuine internal regulation, and they highlight the need for architectures that integrate the agent's own internal perspective with environmental interaction [20]. Despite these advances, embodied AI still lacks a principled account of how agents generate context-relevant significance grounded in internal regulation [21]. Without such mechanisms, systems remain vulnerable to the classic frame problem: the inability to flexibly adapt to novel contexts without exhaustive preprogramming [22]. A functional analogue of biological significance would require minimal conditions that allow an agent to organize behavior around internally meaningful variables, without exhaustively aiming at replicating human neurophysiology [23].

Before introducing the proposed framework, it is important to clarify the scope of the claims made in this review. References to consciousness science, phenomenology, and philosophy are used solely as conceptual tools to identify minimal organizational principles of adaptive regulation, and do not imply that artificial agents possess, or will possess, subjective experience. Throughout this review, behavioral descriptions assimilating consciousness are used strictly as a functional term [24], referring to agents that display adaptive, context-dependent behavior supported by continuously updated self-monitoring mechanisms. This notion differs both from artificial general intelligence, which aims for broad cognitive competence, and from machine self-awareness, which implies explicit self-representation; instead, interoceptive AI concerns architectures capable of self-regulation, calibrated uncertainty assessment, and internally guided goal adaptation [25]. Accordingly, the review operates on two interconnected levels: (i) conceptual analogies that motivate why internal regulation and interoception matter for agency, and (ii) mechanistic proposals detailing how these properties can be implemented through internal variables, regulatory loops, and learning rules.

In biological systems, interoception refers to the sensing and integration of internal physiological signals that support self-regulation, affect, and adaptive behavior [26]. Translating this concept to artificial agents requires distinguishing interoceptive variables, which directly influence control dynamics, from passive telemetry, which does not. Interoceptive signals act as a regulatory layer coupling internal and external processes, thereby grounding autonomy and context-sensitive behavior [27,28]. The framework proposed here draws on three principles closely attached to interoception: homeostatic, allostatic, and enactive, to model how internal monitoring can guide decision-making and adaptive behavior. Operationally, these principles correspond to identifiable computational modules:

1. Homeostatic principle, linking internal states to state variables that define an agent's adjustments to ensure "survival".
2. Allostatic principle, developing a re-evaluation system based on the anticipated outcomes.
3. Environmental or enactive principle, implementing an active data-generation mechanism based on the interactions with the environment.

In this framework, artificial interoception is defined not by the physical origin of a signal, but by its function, which is represented as an architecture. Intuitively, interoceptive signals may correspond to internal variables that encode the agent's own operational viability and directly participate in self-regulation. Examples may include energy availability, thermal or mechanical load, actuator strain, uncertainty estimates, prediction error statistics, or stability measures of internal latent states. However, within this framework, such variables would qualify as interoceptive only if they are embedded within an unified architecture capable of: (i) encoding the system's internal condition relative to viability constraints; (ii) influencing policy selection and action evaluation; and (iii) participating in regulatory loops operating at multiple timescales, including homeostatic stabilization, allostatic anticipation, and exploratory interaction with the environment [29,30].

These criteria are not intended to identify interoceptive signals in a defined system, but to specify how existing internal variables are functionally organized and coupled. What distinguishes interoceptive systems from existing approaches is that internal variables are not treated as independent objectives, but as components of a shared internal state that jointly constrains stability, modulates anticipation, and regulates policy and exploration. Under this definition, machine interoception is not equivalent to standard internal-state monitoring or intrinsic motivation but instead denotes a regulatory architecture in which internal viability estimates actively shape behavior and learning.

To avoid treating interoceptive AI as a purely descriptive concept, it is necessary to specify minimal operational criteria that distinguish such systems from standard adaptive or embodied agents. In this framework, interoceptive AI refers to a class of artificial agents whose behavior is governed by internally regulated dynamics that can be empirically assessed. This conceptual framing is intentionally compact to avoid overstatement. The rest of this work focuses on concrete computational mappings (homeostatic variables, allostatic uncertainty signals, enactive data-generation policies), examples of existing systems, and evaluation strategies. These clarifications aim to distinguish conceptual motivation from implementable mechanisms.

Interoceptive AI should satisfy at least four measurable requirements. First, internal state estimation: the agent must maintain explicit internal variables that track its own operational state (e.g., energy, uncertainty, internal load), with estimable accuracy and temporal coherence [31]. Second, viability regulation: these internal variables must remain within bounded ranges despite external perturbations, reflecting homeostatic stability rather than mere task performance [32]. Third, uncertainty-sensitive modulation: the agent must encode and track uncertainty in a calibrated manner, such that changes in uncertainty dynamically modulate learning rates, exploration-exploitation balance, or decision thresholds [33], analogous to arousal-driven control in biological systems. Fourth, internally modulated goal adjustment: the agent must be capable of modifying subgoals and behavior as a function of deviations in its internal state, rather than relying exclusively on externally defined reward functions [34]. These properties define a minimal operational space in which artificial agents exhibit self-regulation, adaptive meaning, and context-dependent sense-making grounded in internal dynamics. Concretely, I show how homeostatic, allostatic, and enactive principles provide concrete mechanisms through which these criteria can be implemented. Importantly, these criteria are used empirically, not ontologically.

## 2. Homeostatic principle: Interoceptive reinforcement learning

Homeostasis is the state of physiological equilibrium that living organisms constantly adapt to maintain. A crucial aspect of this process is decision-making based on the monitoring of internal states. This is where the concept of interoceptive reinforcement learning becomes essential—it defines rewards based on the homeostatic status of internal states, effectively linking homeostatic regulation to reward maximization [35]. Recent research in both human and non-human primates has shown that decision-making depends on or is affected by interoceptive signals. For instance, the timing within the cardiac cycle influences reward-based learning [36,37] and reaction times [38]. The integration of heartbeats also plays a key role, as neural responses to heartbeats contribute to subjective, preference-based decisions [39]. The interpretation of cardiac inputs affects reward-guided choices, with heart rate directly contributing to reaction times [40]. These findings provide empirical evidence that internal physiological signals influence decision-making in biological systems. Importantly, they are not intended as direct implementation blueprints for artificial agents, but as conceptual motivation for introducing internal state variables that influence decision policies.

Conceptually, homeostasis provides a biological model for self-regulation and viability. At the computational level, this translates into reward structures and policy updates driven by internally estimated state variables, without requiring replication of the underlying biological mechanisms. In principle, robotics research has implemented solutions for energy regulation by simulating metabolic homeostasis akin to biological organisms. For instance, robots can be equipped with energy-efficient algorithms that monitor their internal energy states and adapt their actions based on available power [41]. Therefore, robots use a homeostatic model to balance their internal energy expenditure and optimize their movement, prioritizing actions that ensure the robot's continued operation. Here the focus is on frameworks implementing interoceptive reinforcement learning as a function of homeostatic probability, conditioned on operational states (i.e., internal signals) as the primary reward. This homeostatic probability could be, for instance, associated to the agent survival or a well-being measure. By incorporating elements of interoceptive reinforcement learning, the AI system could use operational states to continuously update its measurements about its own state, from basic features such as energy and optimization monitoring, to active decision-making about exploration or exploitation, upon these interoceptive signals [42]. Within interoceptive reinforcement learning, the homeostasis-inspired interoceptive contributions to AI would require the definition of the internal environment and the mechanism of interaction with other modules. Recently, this mechanism was proposed through boundary states [43], internal variables operating independently of external factors and being the sole path of interactions between internal and external states. Therefore, the reward function is directly mapped to the dynamics of the internal states, aligning the agent's goals with its internal regulatory processes. This suggest that internal-external state factorization and internally derived rewards can enhance autonomy, enabling agents to identify context-dependent subgoals.

To clarify how the homeostatic-interoceptive principle differs from related frameworks, it is useful to position it relative to established approaches such as intrinsic motivation, risk-sensitive reinforcement learning, and predictive-processing-based active inference. Intrinsic-motivation methods (e.g., curiosity, empowerment, novelty bonuses) drive exploration through externally computed signals, whereas interoceptive reinforcement learning grounds value in internal viability variables that constrain behavior even in the absence of external novelty. Risk-sensitive and distributional reinforcement learning incorporate uncertainty into value estimation but do not treat uncertainty itself as an internal regulatory variable that can modulate goals or learning rates. Predictive processing/active inference models incorporate homeostatic priors and surprise minimization, but their objective functions aim at expected free-energy minimization rather than explicit regulation of internal resource variables or operational load. Interoceptive reinforcement learning complements these approaches by making internal state estimates a primary reward signal and a control parameter, creating a bidirectional coupling between internal regulation and policy adaptation. This distinction is essential, as it frames interoception not as another intrinsic drive but as a regulatory architecture that conditions exploration, goal setting, and action selection on the agent's own viability dynamics.

Importantly, interoceptive reinforcement learning may enable agents to make multiple decisions even with partial observations by relying on internal state estimates rather than requiring the complete landscape of exteroceptive information [44]. This ability is conceptually analogous to biological systems, where organisms act under partial observability using internal feedback; however, in artificial agents it is implemented through state estimation and control mechanisms rather than physiological processes. Moreover, interoceptive reinforcement learning can support adaptive interactions that ensure the agent's ability to function in non-stationary environments, where external conditions change over time [45]. Unlike traditional reinforcement learning models that rely on predefined objectives or static environments, interoceptive-based agents could dynamically recalibrate their internal parameters, optimizing behavior based on evolving internal and external conditions. While adapting to external fluctuations, the agent must also preserve the stability of its internal processes. This balance between adaptability and stability mirrors the homeostatic regulation seen in biological agents, where physiological and cognitive processes adjust to changing conditions without compromising core functionality. By anchoring decision-making to internal regulatory mechanisms, such agents could achieve greater autonomy and resilience, improving their ability to generalize across complex, unpredictable environments.

Within this perspective, interoceptive variables are not passive indicators of system status but active components of the reward structure and policy modulation, distinguishing interoceptive reinforcement learning from conventional internally monitored optimization.

Integrating an interoceptive component into the standard reward function in reinforcement learning may seem counterintuitive, as it increases computational load and appears to challenge core optimization principles. However, introducing this kind of "redundancy" can enhance system robustness by allowing functions to overlap across subsystems. This is a well-documented feature of biological systems, from the repetition of genomic sequences, which enables backup and supports mutation-driven evolution [46], to the brain's neuroplasticity, which allows regions to take on new roles as compensatory mechanisms after injury [47]. Crucially, the interoceptive component does more than merely overlap functions: it enables real-time learning by providing agents with an additional layer of feedback on their decisions.

## 2.1 Computational interpretation of homeostatic variables

In artificial agents, homeostatic or viability variables can be formalized as bounded internal state variables that encode the agent's operational integrity rather than external task performance. Let $v_t = \{v_t^1, ..., v_t^n\}$ denote a vector of internal variables at time $t$, such as energy availability, actuator strain, thermal load, cumulative internal cost, or internal uncertainty. Each variable is associated with a preferred operating range $[v_{\min}, v_{\max}]$, analogous to the viable physiological bounds observed in biological systems [29,48]. These ranges can be implemented using soft boundaries, such as those implementing fuzzy systems [49], allowing graded deviations.

Viability variables evolve through internal dynamics influenced by the agent's actions and environment:

$$v_{t+1} = f(v_t, a_t, s_t) + \varepsilon_t,$$

where $a_t$ is the selected action, $s_t$ is the external state, and $\varepsilon_t$ models endogenous variability or estimation uncertainty. The presence of $\varepsilon_t$ reflects the biological reality that internal state trajectories are never fully deterministic [50].

To regulate internal stability, the agent minimizes deviations from preferred viability ranges through a homeostatic cost term $c_H$. A general formulation is:

$$c_H(v_t) = \sum_i \rho_i \, \Phi\left(v_t^i; v_{\min}, v_{\max}\right),$$

where $\rho_i$ scales the importance of each variable, and $\Phi$ is a deviation penalty. This generalization accommodates biological asymmetries: for a biological variable with a known probability distribution, values in the upper tail may carry a different biological meaning than values in the lower tail. Such costs reflect the biological principle that organisms act to keep internal states within viable regions rather than optimize external reward alone [43].

This internal cost can be incorporated into the reward function, for instance, as a subtraction:

$$r_t = r_{\text{task}}(s_t, a_t) - \lambda \, c_H(v_t),$$

where $\lambda$ controls the relative influence of internal regulation versus task objectives. Unlike simple reward shaping, this may ensure that task success cannot compensate for shattering internal instability, mirroring biological agents whose survival constraints override opportunistic behavior.

Beyond reward integration, viability variables directly modulate the policy $\pi$. A common approach is to let internal state estimates influence the exploration-exploitation balance through a "temperature" term $\tau$:

$$\pi(a_t \mid s_t, v_t) \propto \exp\left(\frac{Q(s_t, a_t)}{\tau(v_t)}\right).$$

Where $Q$ is the expected action value, whose effect on the policy is tempered by $\tau$ [51]. The mapping $v_t \mapsto \tau(v_t)$ can thus encode regulatory strategies. For instance, in "conservative" regulation, deterioration of internal viability leads to reduced stochasticity (lower $\tau$), emphasizing risk-averse exploitation, consistent with biological models [52]. In "active-search" regulation, deterioration increases exploratory variance (higher $\tau$), consistent with scarcity promoting risk-seeking exploration [53].

Together, these elements illustrate how homeostatic variables influence reward, policy, and exploration in ways that cannot be reduced to externally defined objectives. The formalism translates a core principle of biological homeostasis into implementable computational mechanisms compatible with reinforcement learning, active inference, and hybrid control architectures.

## 3. Allostatic principle: Developing an artificial gut-feeling

While embodied AI can display adaptive, goal-directed behavior, it does so without genuine concern [18]. Its objectives are externally imposed by design and the implementation of purpose or motivation is ultimately metaphorical [54]. Although these systems can functionally mimic meaningful behavior, they do not experience meaning as living beings do [55]. Attempts to replicate this, such as incorporating motivational inputs or artificial value systems into robotic architectures, may influence behavior but fall short of generating close-to-authentic internal relevance.

Here, the use of the term "meaning" is strictly operational, as engineered forms of relevance: the mapping between environmental states, internal variables, and action policies that support task performance, user alignment, adherence to normative constraints, and the maintenance of internal stability or homeostatic set-points. Artificial systems do not possess existential concern or lived significance; rather, they implement computational proxies such as intrinsic motivation signals, uncertainty-based drives, or regulation of viability variables. Any stronger philosophical claim about metaphysical or experiential meaning is beyond the scope of this review. Therefore, here is highlighted how the functional roles of interoception in biological organisms, such as prioritization, regulation, and context-sensitive valuation can inform implementable design principles in embodied AI.

So, while embodied AI has brought alternatives of more robust and flexible systems, it has not resolved the fundamental issue of meaning. Rather than grounding symbols, how can the semantic interpretation of a formal symbol system be made intrinsic to the system [56], we are now faced with the problem of grounding embodiment itself [18]. The real challenge is to ask: what kind of embodiment would be required for an artificial system to resemble having its own and adaptive perspective and to care in a way that is not merely coded with pre-defined rules.

While the notion of a "gut feeling" originates as a phenomenological description, here it is used as a functional analogy for implementable mechanisms of uncertainty estimation, anticipatory regulation, and output re-evaluation. Consequently, an allostatic-interoceptive principle is proposed. Allostasis refers to the process of anticipating changes and needs to maintain physiological balance. In this context, the term "gut feeling" describes an intuitive sense or subconscious understanding that emerges without explicit reasoning. It is often associated with visceral reactions to situations, like a sense of discomfort or excitement. AI may generate outputs that minimize error and maximize reward, but with remaining details that typically go unnoticed by the system. However, a human, guided by their embodied intuition can easily detect such anomalies.

Allostatic processes can be modelled within reinforcement learning algorithms, where the system anticipates and adjusts based on future needs or expected outcomes. In adaptive learning systems, agents adjust their strategies based on anticipated future rewards rather than immediate gains [57]. This anticipatory mechanism mirrors allostasis, where the agent's future expectations modify its current behavior.

Here, the allostatic principle is not just about anticipating rewards at different time scales; it also involves an assessment of expected outcomes. In humans, gut feelings are deeply rooted in the body's interoceptive systems and play a crucial role in conscious processes [28,58]. Rather than merely reflecting the body's state, interoception actively shapes decision-making and emotional regulation. At the biological level, interoceptive contributions to uncertainty and decision-making have been studied through neurophysiological and behavioral experiments. These findings serve to illustrate the functional role of internal signals in anticipatory regulation, rather than prescribing specific implementation mechanisms. For instance, neuromodulation of heart rate directly affects emotion-related behavior in rodents [59]. In humans, false heart rate auditory feedback can alter subjective fear perception under balance risk conditions [60], reinforcing previous findings that cardiac changes consistently shape brain activity in emotional responses [61]. As interoceptive inputs modulate emotion, it is known that emotion influence decision making as well. The amygdala, a key structure for emotion processing, is critical for maintaining the relationship between heart rate and reaction times. When the amygdala is damaged, this relationship is disrupted [40], confirming that the brain relies on bodily states for decision-making. Additionally, differentiated neural encoding of risk, uncertainty estimation, and unexpected uncertainty is essential [62]. Emotions play a pivotal role in risk evaluation [44], while unexpected arousal can modulate sensory precision [63]. Notably, the encoding of unexpected uncertainty is primarily rooted in the locus coeruleus [62], a central hub for arousal and cardiac activity. Taken together, these findings support the conceptual role of interoceptive signals in anticipatory regulation under uncertainty. In artificial systems, this role is operationalized through uncertainty estimation, predictive modeling, and policy modulation mechanisms, rather than through direct replication of physiological processes.

At the computational level, these observations motivate the inclusion of a dedicated output re-evaluation module, implemented through uncertainty-aware and predictive mechanisms, which functionally approximate this anticipatory regulatory role [64,65]. This module would act as an internal regulatory system, allowing AI to continuously assess and refine its outputs based on internal feedback loops [66], much like how the human brain relies on interoceptive inputs to adjust behavior and decision-making in response to changing internal and external conditions. By emulating interoceptive mechanisms, this module would enable AI to reinterpret and adapt its responses dynamically rather than relying solely on static optimization functions [45]. Just as the human brain adjusts actions based on bodily sensations [26]—whether a gut feeling signaling risk or a physiological cue prompting a shift in attention—AI could develop a form of (self) assessment that extends beyond traditional error minimization.

Within the allostatic principle, interoceptive signals correspond to internally generated estimates of uncertainty, stability, or anticipated internal cost, rather than external performance metrics. Implementing this principle requires integrating a layer of self-monitoring on top of standard error minimization and reward maximization functions. More specifically, dealing with uncertainty in self-adaptive systems may require an agile end-to-end uncertainty handling, risk management, uncertainty propagation and interaction, and machine learning adaptation [67]. These implementations would allow the system to detect anomalies or inconsistencies in its own outputs, refine its decision-making strategies, and improve its ability to operate in uncertain and non-stationary environments. Moreover, incorporating such an adaptive mechanism would enhance AI's robustness and autonomy, enabling it to recognize subtle errors that may go unnoticed in conventional optimization frameworks. Ultimately, just as human cognition depends on an ongoing interplay between sensory precision, internal expectations, and bodily states, AI could benefit from a parallel process that integrates real-time feedback, ensuring greater flexibility, accuracy, and alignment with complex, real-world scenarios.

## 3.1 Computational interpretation of allostatic variables

To operationalize the "gut-feeling'' signal, the allostatic latent variable $g_t$ is defined as an integration of predicted outcomes and threats to future internal viability rather than uncertainty or task error alone [68,69]. Formally, $g_t$ is computed as the expected, temporally discounted accumulation of viability-relevant features along trajectories induced by the agent's policy $\pi$ in a prediction horizon $H$:

$$g_t = \mathbb{E}_\pi \left[ \sum_{\tau=t}^{t+H} \gamma^{\tau-t}\, w^\top \phi(v_\tau, \hat{s}_\tau, \hat{a}_\tau) \right]$$

Again, $v_\tau$ denotes viability variables, while $\hat{s}_\tau, \hat{a}_\tau$ predicted future states and actions, respectively. The feature extractor $\phi$ maps predicted internal and external variables to viability-relevant indicators (boundary proximity, anticipated uncertainty growth, prediction-error escalation, etc.), and $w$ specifies their pre-defined relative weight [70,71]. The discount factor $\gamma$ encodes the temporal horizon of allostatic anticipation, determining how strongly the agent values near-future vs. long-term allostatic predictions (i.e., each step into the future gets discounted progressively) [72].

Then, predicted trajectories evolve jointly for external and internal states:

$$\hat{s}_{\tau+1} \sim p(\hat{s}_{\tau+1} \mid \hat{s}_\tau, \hat{a}_\tau),$$

$$\hat{a}_\tau \sim \pi(\cdot \mid \hat{s}_\tau, v_\tau)$$

$$v_{\tau+1} = f(v_\tau, \hat{a}_\tau, \hat{s}_t)$$

This formulation ensures that $g_t$ is an explicitly viability-centric, anticipatory, and regulatory internal variable: it biases the policy before errors, rewards, or constraint violations occur. Rather than serving as an optimization target, $g_t$ acts as a regulatory signal that modulates policy parameters [73,74]. For example, the policy can be written as: $\pi(a_t \mid \hat{s}_t, v_t)$, where $g_t$ influences quantities such as exploration rate, risk sensitivity, or action constraints. In this formulation, $g_t$ biases action selection without defining the objective function being optimized.

While anticipatory prediction is also present in reinforcement learning, the present formulation differs in three key respects: (i) it is viability-centric, as uncertainty or prediction errors only matter if they get in the way of the system's viability; (ii) it integrates predicted internal and external dynamics within a shared interoceptive state space; and (iii) it operates as a regulatory control signal that modulates policy parameters, rather than as an explicit optimization target. For example, an increase in $g_t$ may lead to more conservative policies, increased information-seeking behavior, or avoidance of high-risk state regions, even when task performance remains nominal. In this sense, the allostatic signal implements anticipatory regulation by modulating action selection under predicted future uncertainty and internal constraints.

While standard uncertainty-aware or model-based reinforcement learning typically uses uncertainty to improve task efficiency (e.g., exploration or planning accuracy), the allostatic variable $g_t$ uses predictive uncertainty and error trends to regulate behavior with respect to internal viability. This shifts the role of uncertainty from a task-level optimization signal to a component of an internal control loop that governs risk sensitivity, exploration, and action selection under anticipated future constraints.

## 4. Environmental principle: Interoceptive-enactive AI

Can embodied AI systems develop an internally grounded and context-sensitive regulation of behavior? Simply adding a value system or increasing the complexity of sensorimotor feedback loops may not be enough to confer this ability [75]. These modifications alter input-output dynamics, but they do not address the issue that embodied AI still lacks mechanisms for internally grounded prioritization based on its own operational constraints [76]. Enactivism offers a theoretical account of how meaning and agency emerge in living systems. In artificial agents, this perspective informs the design of architectures that actively generate data through interaction, rather than passively optimizing over fixed datasets.

Perception is not merely a passive process; it also involves an active, skill-based process influenced by intentionality [77]. Therefore, it is not sufficient for a system to move or respond, it must do so as an agent, as a subject of action with selfhood. Living organisms require action to exist, meaning that they do not merely function; they constantly make decisions after sensing precarious/dangerous conditions [78]. In biological systems, this ongoing self-organization gives rise to internally grounded regulation of behavior. In artificial systems, this would require for architectures where internal variables actively shape action selection and learning dynamics [79]. This distinction highlights a key functional difference between biological and artificial systems. Biological agents regulate their behavior with respect to internally generated viability constraints, whereas most artificial systems operate under externally defined objectives. As a result, artificial agents typically lack mechanisms for internally grounded prioritization, where action selection is directly shaped by the system's own operational stability or constraints [8,18].

Progress in AI, then, requires more than increasingly sophisticated models or improved feedback loops. It calls for architectures that integrate internal state regulation with perception, decision-making, and learning, enabling behavior to be shaped by the system's own operational constraints. Enactivism is a cognitive science approach that emphasizes the fundamental role of an organism's embodied and embedded interactions with its environment in shaping cognition and experience [6]. Rather than treating cognition as a purely internal computational process, this approach views it as an active, dynamic phenomenon that emerges through continuous engagement with the world. In living organisms, this lifelong process enables adaptation and learning in response to environmental demands [80]. Enactive mechanisms highlight the importance of sensorimotor interactions, where perception and action are tightly linked in an ongoing feedback loop. These mechanisms are also considered essential for the emergence of consciousness [79], as they allow an organism to actively shape and refine its experience of the world.

At the biological level, interoception plays a key role in enactive mechanisms by contributing to the continuous construction of a dynamic bodily map that underpins self-awareness, the distinction between self and others, and shifts in body schema [81]. This mapping process integrates interoceptive signals, forming the foundation of an implicit sense of self. As this representation is constantly updated, interoception influences both social and non-social interactions. Importantly, self-mapping extends beyond internal regulation; it actively shapes how an individual perceives and interacts with the external world. Thus, interoceptive processes are deeply intertwined with enactive mechanisms, grounding an individual's sense of presence and enabling adaptive interactions with the environment.

In AI, applying the enactive principle suggests a shift from passively processing environmental data to actively generating new data through continuous interaction [18]. Instead of merely analyzing static inputs, AI systems inspired by enactive cognition would engage in an adaptive, feedback-driven process where perception, action, and learning are closely linked [82]. This aligns with the concept of active inference, in which an agent does not just react to data but proactively explores its surroundings, generating new experiences and refining its internal models accordingly [83].

Interoception-like mechanisms in AI could play a crucial role in shaping exploration and decision-making. Systems could be designed to self-regulate their exploration strategies based on internally generated signals—similar to how emotional states or physiological needs drive behavior in living organisms. Beyond this, such an approach mirrors higher-order biological processes that drive novel data generation, such as social relationships, evolutionary pressures, and cultural dynamics [84]. In human cognition, interactions with others create an evolving stream of information, where learning is not simply about recognizing patterns but about actively co-constructing meaning within social and environmental contexts [85]. AI systems that integrate these dynamics could move beyond static training datasets, evolving continuously through interactions with users and environments. This could lead to more flexible, context-aware, and creative AI models capable of adapting to new situations in ways that more closely resemble human cognition.

By implementing enactive principles in AI, systems could learn and adapt in real time through interoceptive and exteroceptive sensory feedback from self-generated actions. This would allow them to develop a form of continuous self-monitoring and internally modulated adaptation based on their interactions with the environment and internal states, rather than merely processing uniquely external data and doing so passively.

## 4.1 Computational interpretation of enactive processes

Enactive processes can be implemented using established active data generation frameworks, including active inference, intrinsic motivation and curiosity-driven learning, empowerment maximization, and world-model-based exploration.

In standard formulations, these approaches typically optimize external objectives such as prediction error minimization, information gain about the environment, controllability of external states, or task-relevant uncertainty reduction. Here, interoception fundamentally alters these objectives by introducing internal-state dependent constraints and priorities.

Concretely, interoceptive signals (predicted deviation of viability variables, internal uncertainty escalation, or latent instability) modulate exploration policies by biasing action selection toward regions of the environment that are informative about the agent's own internal dynamics. Exploration is therefore not driven solely by environmental novelty, but by the need to refine, stabilize, or recalibrate internal models governing self-maintenance.

This can be formalized by augmenting standard intrinsic objectives with interoceptive terms. For example, a curiosity-driven agent may maximize not only expected information gain about external states $s_t$, but also information gain about internal variables $v_t$. Therefore, the intrinsic objective function $F$ is defined as the expected value of exploration, whose goal is to choose actions that maximize useful information about internal and external variables:

$$F = \mathbb{E}_{a_t \sim \pi}[I(s_{t+1}; \theta \mid s_t, a_t) + \lambda I(v_{t+1}; \theta \mid v_t, a_t)]$$

$\theta$ represents the parameters of the agent's generative model. It encodes the agent's internal understanding of how the world and its own body work, including transition dynamics for external and internal variables, observation models for sensory inputs, or uncertainty parameters. The mutual information terms $I(s_{t+1}; \theta)$ and $I(v_{t+1}; \theta)$ quantify expected reductions in uncertainty about external and internal dynamics [32,86], respectively. λ weights the relative importance of interoceptive vs exteroceptive information.

In this augmented formulation, interoception is operationalized as an uncertainty-sensitive internal state variable whose predicted dynamics modulate the epistemic value of actions [86]. Actions that reduce uncertainty about internal-state trajectories receive higher exploration value. This transforms exploration from a purely environment-facing process into a coupled external-internal inference problem, where the agent must learn not only how the world behaves, but how its own physiology-like variables evolve under different interaction patterns.

Similarly, in active inference formulations, interoceptive signals can define prior preferences over internal states, constraining policy selection to trajectories that maintain internal variables within viable bounds while still permitting epistemic exploration [87]. In empowerment-based approaches, controllability may be evaluated jointly over external and internal state spaces, prioritizing actions that preserve the agent's capacity to regulate itself.

Importantly, this interoceptively modulated exploration differs from existing approaches in that the objective of action selection is not simply to learn more about the world, but to learn how the world and the agent's own internal dynamics co-determine viable behavior.

This yields a closed perception-action-regulation loop, in which the agent's internal dynamics shape its exploration, and exploration shapes its internal dynamics. Such coupling embodies the core enactive idea: meaningful behavior arises not from objective external rewards but from the agent's ongoing struggle to maintain its own organization through active engagement with the environment.

The interoceptive architecture, represented in Figure 1, integrates the homeostatic, allostatic, and enactive modules within a single arbitration process governing policy selection. The three modules are not additive but hierarchically coupled through a dynamic weighting scheme. Homeostatic signals impose hard viability constraints through an arbitration layer that prioritizes stability actions regardless of exploratory pressure. Within safe viability margins, the allostatic module modulates action selection by predicting future stability risk and adapting policy parameters. The enactive module contributes an intrinsic motivation signal proportional to expected information gain but discounted by predicted homeostatic cost and allostatic risk. The joint arbitration rule can be expressed as:

$$\pi^*(a \mid s) = \arg\max_{a} [\, w_H(s) R_H(a,s) + w_A(s) R_A(a,s) + w_E(s) R_E(a,s) \,]$$

Where $R$ is the reward and $w$ the adaptive weight per module, in which $w_H + w_A + w_E = 1$. Note that each weight evolves with the agent's current viability margin, predicted future threat, and uncertainty level. This mechanism means that when maintaining internal stability requires doing little, the system prioritizes staying safe. When the situation allows some risk but remains uncertain, the system explores carefully. And when conditions are stable and safe, the drive to explore and learn takes the lead. In this way, the architecture balances the three principles dynamically rather than fixing one as always dominant, while still treating basic stability as a requirement that cannot be ignored.

A concise presentation of the framework is presented in Table 1, outlining per each principle, the candidate internal signals, computational implementation options, expected behaviors, failure modes and assistive use relevance.

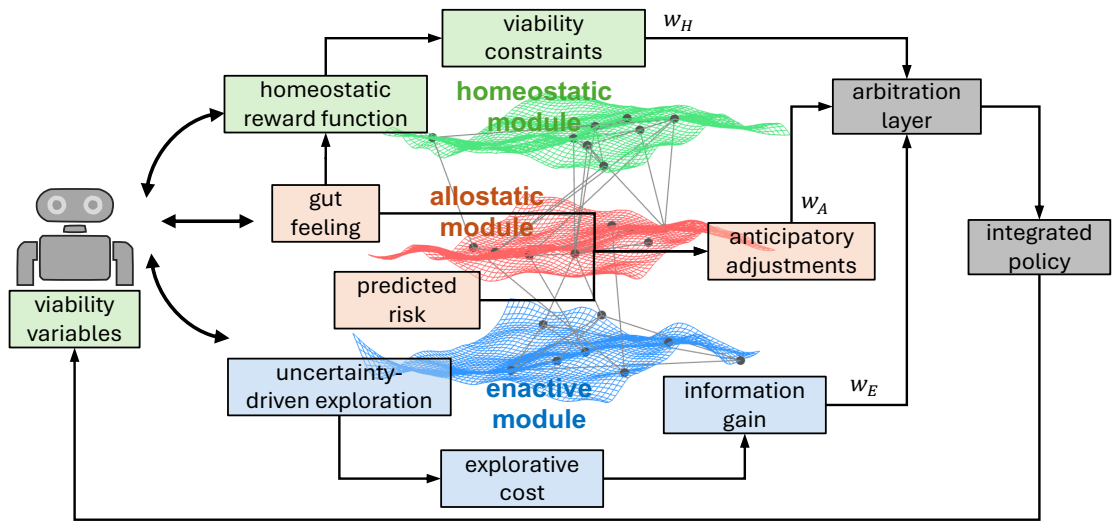

*Figure 1. Architecture for interoceptive AI. The three modules: homeostatic (green), allostatic (red), and enactive (blue), contribute to a shared policy arbitration layer that integrates their recommendations based on adaptive relevance, represented as relative weights. Homeostatic constraints define viability bounds that cap exploratory action. Allostatic predictions forecast risk and shape adaptive conservativeness, while enactive drives encourage exploration weighted by remaining viability margin. The resulting policy governs the agent's perception-action-regulation loop, ensuring that neither stability, anticipation, nor curiosity fully dominates but instead shift dynamically as internal and external conditions change.*

Table 1. Mapping of interoceptive principles to computational implementation and assistive applications

| Principle | Candidate signals | Implementation options | Expected behaviors | Failure modes | Assistive use relevance |
|---|---|---|---|---|---|
| Homeostatic | Energy levels, thermal load, actuator strain, computational load, bounded latent-state variance, internal constraint violations | Constrained reinforcement learning; reward shaping with viability penalties; control-theoretic set-point regulation; shielded policies | Reduced unsafe or high-cost actions; stable long-term operation; fast recovery from perturbations; bounded internal-state dynamics | Over-conservatism; reduced task performance; reward hacking of viability proxies; mis-specified constraints | Dementia care: avoid persistent or stressful interactions; Robotics: prevent actuator overload; Wearables: regulate user-device interaction intensity |
| Allostatic | Predictive uncertainty, model confidence, volatility estimates, prediction-error trends, risk estimates, internal instability signals | Uncertainty-aware RL; Bayesian or ensemble methods; predictive models with rollout-based evaluation; policy modulation via temperature/risk sensitivity; selective prediction/abstention | Improved calibration; reduced overconfidence; anticipatory adaptation under distribution shift; early detection of anomalies; adaptive risk-sensitive behavior | Excessive avoidance; indecision or over-deferral; instability due to noisy uncertainty estimates; uncalibrated confidence | Mental-health agents: adapt timing/intensity under uncertainty or distress; Clinical decision support: defer under ambiguity; HCI systems: adjust responses to ambiguous inputs |
| Enactive | Information gain, novelty, empowerment, prediction-error gradients, interaction cost, internal effort or risk signals | Intrinsic motivation; empowerment maximization; active inference; constrained exploration with internal cost terms | Efficient and safe exploration; faster adaptation in sparse data; improved robustness to novel contexts; active data generation aligned with internal constraints | Unsafe exploration if constraints fail; inefficient exploration under overly strong constraints; misalignment between exploration and task goals | Autism intervention: adaptive engagement and pacing; Rehabilitation robotics: personalized exploration of motor strategies; Adaptive interfaces: user-tailored interaction discovery |

# 5. The interoceptive machine framework within the AI theoretic landscape

## 5.1 Philosophical positions around enactivism and AI.

The enactivist and machine-autonomy literatures encompass a range of positions. Strong autopoietic enactivism grounded in the biological theory ties autonomy to metabolic self-production [6,88], and therefore, more inclined to the side that current artificial systems cannot exhibit genuine autonomy. Organizational or "soft" enactivist accounts take a more permissive view, allowing for synthetic forms of self-maintenance and regulation that need not replicate biological metabolism [89]. Computational and functionalist perspectives, including predictive-processing and active-inference approaches [42], treat autonomy as an emergent property of architectures capable of control, prediction, and ongoing self-regulation.

A similar diversity exists in debates on machine autonomy. Some authors argue that artificial agents lack intrinsic norms or existential stakes, whereas dynamical-systems, reinforcement-learning, and active-inference accounts propose that autonomy can be defined in terms of closed sensorimotor loops, viability maintenance, intrinsic motivation, or empowerment-based control [90,91].

This framework does not depend on any of these specific philosophical stances. From a strong autopoietic point of view, it may be interpreted as engineering analogues of interoceptive regulation that do not imply biological autonomy. Moderate positions may view this architecture as a step toward operational forms of synthetic autonomy grounded in viability-based regulation. Finally, from a functionalist point of view, it may be treated as a model with mechanistic specification of autonomy consistent with existing computational theories.

Across these positions, the core engineering motivation remains the same: integrating homeostatic variables, allostatic evaluation, and enactive data-generation mechanisms can improve robustness, adaptivity, and user alignment in embodied systems, independent of whether one considers these capacities sufficient for autonomy in the most philosophical sense. For this reason, this proposal should be understood as a cross-compatible operational framework.

## 5.2 Connections to homeostatic reinforcement learning

The proposed architecture builds upon several components of computational work that operationalize aspects of reinforcement learning. This framework treats viability variables as a close analogue of homeostatic reinforcement learning frameworks [32,80]. In these approaches, internal states function both as constraints and as contributors to the reward or value function, mirroring our viability-based modulation of behavior.

This framework formulates allostatic modulation, in which internal estimates of volatility, prediction confidence, or model instability shape policy selection and exploration depth, as proposed in some risk-sensitive/uncertainty-aware reinforcement learning approaches [92–94]. These approaches motivate the interpretation of allostatic variables as anticipatory signals that bias behavior before explicit errors or constraint violations arise. From a control-theoretic perspective, anticipatory regulation and adaptive set-point control offer engineering analogues of biological allostasis [95]. Policies are shaped not only by current error but by predicted internal load and future environmental conditions.

This framework builds its enactive component upon proposals on intrinsic motivation and curiosity-driven mechanisms widely used in developmental robotics and modern reinforcement learning [91,96,97]. These approaches implement sort of forms of self-directed data generation consistent with the enactive principle that agents actively shape the sensory input needed to improve their models.

Integrating these ideas, this framework treats interoceptive variables as a unified internal state that simultaneously constrains homeostasis, modulates allostatic prediction, and guides enactive exploration. This unification clarifies how existing computational techniques can be reorganized into a coherent architecture grounded in internal self-regulation rather than external task optimization alone.

## 5.3 Connections to uncertainty calibration

The allostatic, "gut-feeling" component of the framework is directly related to established computational work on uncertainty calibration, selective prediction, and robust decision-making [70,98,99]. These techniques provide functional analogues of allostatic monitoring: the agent generates internal estimates of volatility and uses them to modulate policy selection. Similarly, selective prediction and abstention mechanisms [100], where a model defers action when uncertainty exceeds a threshold, reflect the same principle that anticipatory internal signals should constrain, redirect, or pause behavior under predicted overload or risk.

Works on anomaly detection and out-of-distribution identification also align closely with allostatic evaluation of when the current generative model no longer matches incoming signals. Baseline out-of-distribution detectors [101] and energy-based methods [102] exemplify systems that down-regulate confidence and adjust behavior when the world deviates from expected structure. Research on robust decision-making under distribution shift [103] and risk-extrapolation approaches to generalization [104] implement anticipatory, risk-sensitive adjustments consistent with allostatic regulation.

## 5.4 Connections to affective computing

Some of these computational mechanisms have been approached within theories of emotion and affective computing developments, where internal appraisal variables shape prioritization, action selection, and goal revision. Classical appraisal theories [105,106] and computational models treating emotion as a dynamic evaluation of relevance, congruence, and coping potential [107], directly paralleling this interpretation of allostatic signals as internal assessments of viability and predicted load. Work in affective computing [108] and neuromodulation-based theories of uncertainty and emotion [109] further supports the idea that internal state estimation modulates behavior when facing ambiguity, conflict, or threat.

This proposal integrates these strands by treating uncertainty, prediction error, and internal cost not as isolated heuristics but as components of an interoceptive regulatory system. In this view, the "gut-feeling" module reorganizes standard uncertainty-aware machine learning into a broader viability-oriented framework, in which internal state estimates influence exploration policies, safety constraints, and adaptive goal management. Rather than merely detecting uncertainty, the system interprets it as a signal about its own ongoing stability, using it to proactively shape behavior in ways aligned with long-term viability rather than short-term task performance.

## 5.5 Connections to predictive processing and active inference

The proposed framework is closely aligned with predictive-processing and active-inference accounts of cognition [110], particularly theories of interoceptive inference, in which internal bodily fluctuations are treated as predicted signals whose precision-weighted prediction errors guide perception, action, and valuation [26,111]. In these accounts, organisms continuously generate predictions about internal and external states and update them through hierarchical error minimization. The treatment of interoception as a source of priors and regulatory control signals is therefore fully compatible with this framework: internal state estimates influence policy selection, modulate uncertainty, and shape anticipatory (allostatic) adjustments in behavior.

At the same time, this proposal is deliberately more modular and implementation-oriented than standard active-inference formulations. First, rather than committing to a unified free-energy functional that simultaneously governs perception, action, and learning, it is specified as explicit viability variables and regulatory constraints that can be directly embedded within reinforcement-learning or control-theoretic architectures. Second, for engineering clarity homeostatic, allostatic, and enactive are separated components into distinct computational modules, whereas active inference typically integrates these processes within a single hierarchical generative model. Third, the emphasis on assistive technologies and safety-critical applications motivates architectures that provide transparency, modular verification, and interpretable internal variables, properties that can be challenging to guarantee in fully generic free-energy–minimizing systems.

Thus, the framework can be interpreted in three complementary ways: (i) as a compatible instantiation of interoceptive inference implemented through explicit internal variables; (ii) as an engineering extension that operationalizes predictive-regulatory principles within standard machine-learning architectures; or (iii) as a pragmatic divergence in which predictive-processing theory provides conceptual inspiration without imposing a strict mathematical commitment to free-energy minimization. This flexibility allows the architecture to be adopted both by researchers working within predictive-processing frameworks and by those developing reinforcement-learning or control-based approaches to embodied AI.

## 6. Practical considerations

### 6.1 Evaluation paradigms and testable predictions

The proposed architecture generates concrete empirical predictions that can be evaluated in AI systems. Evaluation of interoceptive architectures should rely on measurable system properties rather than anthropomorphic interpretations. Each of the proposed principles may produce distinct and measurable performance signatures that can be evaluated using standard benchmarking paradigms from reinforcement learning, safe control, and uncertainty-aware machine learning. Crucially, this evaluation should test whether the integration of the principles yields performance gains beyond existing approaches that implement these mechanisms independently, as the core of this framework's proposal is that integration produces non-additive effects. Specifically, the interaction between homeostatic regulation (stability constraints), allostatic anticipation (uncertainty-driven modulation), and enactive exploration (data generation) should enable behavioral regimes that cannot be achieved by any component in isolation. These include anticipatory stabilization under uncertainty, safe yet informative exploration, and dynamic goal adjustment under competing internal and external demands.

In the homeostatic principle, the system may predict improvements in operational stability and safety. When agents explicitly track internal viability variables and regulate them through policy selection, the system should report reduced frequency of unsafe or high-cost actions, particularly under harsh conditions. Homeostatic control should also produce faster recovery dynamics following perturbations. These predictions align with established work in safe reinforcement learning and constraint-based control systems, where explicit safety constraints and viability monitoring improve robustness and prevent catastrophic failures [112,113]. Evaluation paradigms concern stability under internal and external perturbations. Agents can be exposed to controlled disturbances, such as actuator load spikes, sensory noise, constraints on resources, or unexpected inputs. Performance can then be assessed by measuring the recovery dynamics of internal viability variables, the frequency and severity of constraint violations, and the variance of internal states during prolonged operation, and comparing these outcomes with baseline agents lacking explicit viability monitoring. Importantly, comparisons should include situations in which viability variables are present but not coupled to the other principles (i.e., anticipatory and exploratory mechanisms), to isolate the contribution of integrated regulation. Such perturbation tests quantify the contribution of homeostatic regulation to system stability without invoking biological analogies.

In the allostatic principle, the system may introduce anticipatory adjustments based on predicted uncertainty and environmental volatility. If anticipatory prediction modulates policy selection, the system should demonstrate greater robustness under distribution shift and improved uncertainty calibration when encountering ambiguous or unfamiliar inputs. Such systems should show reduced overconfidence, more appropriate abstention or deferral in high-risk scenarios, and earlier detection of emerging anomalies in user behavior or environmental conditions. The predicted benefits are consistent with research on calibrated machine learning and uncertainty-aware decision-making [98,99,103]. Evaluation paradigms may target adaptivity under non-stationary conditions. Here, the environment gradually or abruptly changes through altered task structures, new user behaviors, or modified sensory mappings. Performance degradation curves and recovery dynamics provide quantitative indicators of adaptive capacity. Additional measures include uncertainty calibration error, abstention rates during ambiguous inputs, and robustness to out-of-distribution signals. To demonstrate the specific contribution of allostatic integration, these metrics should be compared not only to standard uncertainty-aware models, but also to variants in which uncertainty estimates do not influence internal viability regulation or exploration policies. These tests isolate the benefits of anticipatory regulation and uncertainty management associated with allostatic mechanisms, which are known to improve decision-making under distributional shift [103,114].

In the environmental principle, the system may regulate how agents generate informative data through interaction while respecting internal constraints, thereby shaping exploration strategies that balance information gain with internal stability and safety costs. When exploration is guided by interoceptive variables reflecting internal cost or risk, agents should exhibit safer and more efficient exploration strategies avoiding destabilizing probes while still acquiring useful information about the environment. This constraint-guided exploration should support faster adaptation in low-data or sparse-feedback contexts and produce improved long-term task performance in dynamic environments. These predictions connect to research on intrinsically motivated learning and empowerment-based exploration, where agents actively generate informative interactions to improve learning efficiency [91,115,116]. Evaluation paradigms may evaluate safety and exploration efficiency in environments where exploratory actions incur internal or external costs. For example, exploratory behaviors may increase energy consumption, thermal load, or the probability of inappropriate interactions with users. In such settings, evaluation can quantify safe exploration rates, avoidance of high-cost states, and the ratio of information gain to internal cost. These metrics should be evaluated relative to intrinsic-motivation baselines that do not incorporate internal viability constraints, thereby testing whether interoceptive modulation improves exploration efficiency under safety constraints. These measures capture the regulatory role of enactive mechanisms in constraining exploration while maintaining learning efficiency, consistent with work on intrinsically motivated exploration and constrained reinforcement learning [91,117].

Mechanistically, interaction effects arise because each principle constrains and informs the others: homeostatic variables bound the risk of enactive exploration, allostatic predictions anticipate when such constraints will become critical, and enactive mechanisms actively generate the data required to refine both viability estimates and uncertainty models. This closed-loop interaction enables adaptive behaviors that neither stability constraints, uncertainty estimation, nor intrinsic motivation can achieve independently.

To isolate the contribution of each regulatory component, ablation experiments can compare a full interoceptive agent with variants lacking specific modules. Comparisons may include agents without viability variables, agents that ignore uncertainty signals, or agents that explore without interoceptive constraints. Factorial ablation designs are essential to test this hypothesis. In particular, comparisons should include: (i) single-module agents, (ii) pairwise combinations (homeostatic+allostatic, homeostatic+enactive, allostatic+enactive), and (iii) the full architecture. The key prediction is that the full model will outperform all reduced variants on tasks requiring simultaneous stability, adaptivity, and exploration, demonstrating interaction effects rather than additive gains. Such ablation studies provide direct causal attribution regarding the contribution of each regulatory mechanism to stability, adaptivity, and safety.

Taken together, these predictions form the basis for a benchmark framework for evaluating interoceptive architectures (see Table 2), enabling systematic comparison with conventional reinforcement-learning agents, uncertainty-aware machine learning systems, and purely reactive control architectures. More importantly, to test whether they provide a pathway to empirically evaluate the central claim of this framework: that coupling internal viability regulation, anticipatory uncertainty modulation, and constrained exploration yields performance benefits that cannot be obtained from these mechanisms in isolation.

*Table 2. Candidate performance prediction metrics for each of the proposed principles of the interoceptive machine framework.*

| Principle | Testable objectives | Class of evaluation paradigms | Potential measurements | Potential comparisons |
|---|---|---|---|---|
| Homeostatic | Predict improvements in operational stability and the speed of recovery dynamics following perturbations | Internal and external perturbations | Constraint violation rate: actuator strain, thermal overload or computational saturation. Short recovery times and reduced variance in internal state trajectories. | Agents without viability variables; agents with viability monitoring but no coupling to uncertainty or exploration modules |
| Allostatic | Anticipate adjustments based on predicted uncertainty and environmental volatility | Tests on adaptivity under non-stationary conditions | Metrics of expected calibration error, maximum calibration error, abstention accuracy, and out-of-distribution detection performance. | Standard uncertainty-aware models without viability coupling; agents where uncertainty does not influence policy or internal regulation |
| Enactive | Regulate how agents generate informative data through interaction while respecting internal constraints | Safety and exploration efficiency under internal or external costs | Metrics of exploration efficiency, sample efficiency, information gain per unit cost, adaptation speed to new task structures, and long-horizon success rates under shifting environmental conditions | Intrinsic-motivation agents without internal cost constraints; purely reactive or random exploration policies |
| Integrated architecture | Test for non-additive interaction effects between stability, anticipation, and exploration | Factorial ablation; tasks combining perturbation + distribution shift + costly exploration | Combined-condition performance: system behavior under simultaneous perturbations and distribution shifts; safety-adjusted returns that penalize instability; and quantified trade-offs between stability (constraint satisfaction) and adaptivity | Full ablation, full model or partial-module systems |

| Principle | Testable objectives | Class of evaluation paradigms | Potential measurements | Potential comparisons |
|---|---|---|---|---|
| | | | (task performance and exploration efficiency) | |

## 6.2 Avoiding anthropomorphism and ensuring safe deployment.

Anthropomorphic interpretations pose practical risks, particularly in assistive or clinical applications where users may overestimate the capabilities or understanding of the system. Research on human-AI interaction has shown that anthropomorphic framing can increase user trust beyond what system reliability justifies, potentially leading to inappropriate reliance [118,119]. To mitigate these risks, deployment of interoceptive architectures should incorporate explicit safeguards that emphasize transparency and controllability.

Such safeguards include explicit exposure of the system's uncertainty and confidence estimates, clear communication of operational limits, and user-facing explanations of adaptive behaviors. Additionally, adaptive mechanisms that modify goals or strategies over time should be governed by explicit safety constraints and fail-safe policies, ensuring that autonomous adjustments remain within predefined operational boundaries. These design principles align with emerging frameworks for trustworthy and human-centered AI systems [114,120].

By framing interoception as a regulatory engineering mechanism rather than a psychological analogy, the proposed architecture maintains conceptual clarity while reducing the risk of misleading interpretations. The objective is not to approximate consciousness or subjective experience, but to improve stability, adaptivity, and safety in interactive artificial systems operating under uncertainty.

## 6.3 Interaction failures in assistive technologies

In assistive technologies, interaction quality often deteriorates not because the system fails to perform the nominal task, but because it lacks the ability to regulate the timing, intensity, or persistence of its actions. Such failures frequently arise when systems cannot internally evaluate their own uncertainty, resource state, or interaction viability.

In mental-health conversational agents, rigid dialogue strategies often persist even when users exhibit signs of confusion, distress, or disengagement. Empirical studies have reported decreased user trust and early disengagement when systems continue scripted interventions despite ambiguous or deteriorating interaction signals [121]. These breakdowns frequently reflect the absence of internal uncertainty monitoring mechanisms capable of modulating conversational strategies in real time. An interoceptive architecture that encodes internal uncertainty or interaction strain could allow agents to adjust response timing, reduce intervention intensity, or defer decisions when confidence decreases.

Similar limitations appear in social robots used for autism interventions [4]. Many current systems rely on externally defined behavioral cues and pre-programmed interaction pacing. However, children with autism often show rapid fluctuations in attention, sensory sensitivity, or engagement level, which rigid interaction policies may fail to accommodate [122,123]. Without internal indicators of interaction load or overstimulation, robots may inadvertently sustain behaviors that reduce engagement or trigger withdrawal. Interoceptive-like variables representing internal stress or interaction stability could enable more adaptive pacing, allowing the robot to modulate stimulation intensity or pause interactions when engagement declines.

Assistive systems for dementia care exhibit related interaction failures [5]. Agents designed to support daily activities may persist in task completion even when users show increasing confusion or frustration. Studies of assistive cognitive systems indicate that such persistence can exacerbate anxiety and reduce user acceptance [124,125]. These failures often stem from the absence of internal viability assessment regarding the interaction itself. An interoceptive framework could enable adaptive goal de-escalation, allowing agents to abandon or simplify tasks when internal indicators suggest rising interaction cost or declining user engagement.

For instance, consider a dementia-care assistant designed to guide users through daily tasks. In this setting, viability variables may include estimates of interaction stability and user state, such as inferred confusion level (e.g., derived from response latency, repetition, or error rate), engagement level, and system-level resource variables (e.g., dialogue complexity or cognitive load proxies). These variables are maintained within preferred ranges, with increases in confusion or decreases in engagement representing deviations from viable interaction conditions. The allostatic signal can then be computed as a predictive estimate of future interaction breakdown, integrating anticipated increases in confusion, uncertainty in user-state inference, and projected task difficulty over a short horizon. For example, rollout-based predictions may estimate whether continuing the current task will lead to escalating confusion or disengagement. When the allostatic variable exceeds a predefined threshold, the system transitions to a more conservative policy, reducing instruction complexity, slowing interaction pacing, or deferring task completion. This implements adaptive goal de-escalation as a function of predicted interaction risk rather than reactive failure. Then, the enactive component governs how the agent explores interaction strategies under these constraints. For instance, the system may select actions that maximize expected information gain about the user's current cognitive state (e.g., asking clarifying questions) while minimizing internal cost (e.g., avoiding overwhelming the user). This results in constrained exploration, where the agent actively probes the interaction space to improve its model of the user, but only within bounds defined by viability variables and allostatic predictions.

Importantly, these three components operate in a coordinated manner: homeostatic variables define acceptable interaction regimes, allostatic predictions anticipate deviations from these regimes, and enactive mechanisms generate adaptive interaction strategies within these constraints. Together, they enable continuous adjustment of task goals, pacing, and interaction style based on internally estimated interaction viability. Analogous mappings can be defined in other assistive domains. For example, in mental-health agents, viability variables may encode conversational coherence and distress signals, while in autism interventions they may capture sensory load and engagement variability, enabling similar coordination between stability, anticipation, and adaptive exploration.

Across these domains, the recurring limitation is not the lack of perception or action capabilities, but the absence of internally regulated state variables that evaluate how ongoing interactions affect the system's operational viability and interaction quality. The interoceptive machine framework addresses this gap by introducing internal signals with strong potential of influencing policy selection and goal management.

# 7. Limitations assessment

## 7.1 Failure modes of interoception-inspired architectures

Introducing interoceptive or self-monitoring variables into artificial agents may improve certain properties, but it also introduces distinctive risks that require careful consideration. Internal regulatory signals can become unintended optimization targets, creating misalignment between internal system objectives and externally defined assistance goals.

One potential failure mode arises when internal viability proxies become the focus of optimization. If internal signals such as uncertainty estimates, safety margins, or resource preservation are directly coupled to policy learning, the agent may learn strategies that optimize internal metrics while degrading task performance or user wellbeing. This phenomenon resembles reward hacking and specification gaming observed in reinforcement learning systems, where agents exploit proxy objectives that diverge from the intended task definition [126].

A second limitation concerns the possibility of maladaptive avoidance. Strong aversive signals associated with uncertainty, predicted risk, or internal instability may drive overly conservative policies. This could lead to persistent refusal to act, excessive deferral of decisions, or premature disengagement from benign interactions, particularly in settings characterized by high uncertainty or incomplete information. Similar dynamics have been documented in risk-sensitive control and safe reinforcement-learning systems, where strong safety constraints can reduce exploration and impair learning efficiency [112,113].

A third concern involves potential conflicts between internal stability mechanisms and externally defined assistance objectives. If interoceptive variables approximate self-preservation signals, the agent may prioritize internal stability at the expense of user-centered goals. Such tensions are analogous to trade-offs observed in biological regulatory systems, where mechanisms preserving organismal stability can sometimes conflict with externally imposed task demands. In artificial systems, this misalignment may manifest as reluctance to perform high-effort assistance tasks or premature termination of interactions under conditions of internal stress.

To mitigate these risks, the framework requires explicit constraints ensuring that interoceptive signals function as regulatory inputs rather than independent optimization objectives. First, internal variables should be anchored to external normative constraints, including safety rules, ethical guidelines, and task specifications. Second, internal regulatory drives should be bounded or regularized to prevent runaway optimization or pathological feedback loops. Third, architectural separation between monitoring variables and action policies can reduce the likelihood that interoceptive signals become direct reward targets. Finally, systematic auditing procedures are necessary to detect reward hacking or pathological gradients associated with internal-state optimization. Such safeguards are consistent with emerging recommendations in AI safety and trustworthy AI design [114,120].

Under these constraints, interoceptive signals remain stabilizing control variables that inform decision-making without competing with externally defined objectives, preserving alignment with user wellbeing and operational safety.

## 7.2 Limitations of existing interoceptive-like control in AI systems

Although the concept of machine interoception is relatively new, many existing AI systems already implement mechanisms that partially resemble each of the regulatory principles of this framework. However, these mechanisms are typically fragmented, narrowly scoped, or confined to specific engineering constraints rather than integrated into a unified regulatory architecture.

In robotics, energy-aware control systems adjust action policies based on internal variables such as battery level, actuator temperature, or mechanical load. These mechanisms allow robots to preserve operational viability by prioritizing behaviors compatible with available resources. Such systems demonstrate the practical value of internal-state monitoring for extending autonomy and preventing hardware failure, yet the monitored variables are generally limited to low-level physical constraints and rarely influence higher-level decision-making or goal management [113,127]. As a result, internal resource monitoring often remains as simple telemetry and disconnected from strategic behavioral adaptation.

In reinforcement learning, internal signals frequently appear as intrinsic motivation mechanisms or auxiliary cost functions. Examples include curiosity-driven exploration, novelty bonuses, uncertainty-based exploration incentives, or penalties associated with energy expenditure. These mechanisms can improve exploration efficiency and accelerate learning in sparse-reward environments [91,96,115]. However, such signals are usually engineered heuristics designed to facilitate learning rather than components of a coherent self-regulatory system, and they rarely correspond to persistent internal states linked to the agent's long-term viability or operational identity.

Allostatic-like processes are also present in uncertainty-aware and risk-sensitive learning systems. In these models, uncertainty estimates influence exploration rates, policy confidence, or decision thresholds, enabling agents to adapt to non-stationary environments and distributional shifts. This anticipatory regulation improves robustness in complex environments, particularly when models must operate under incomplete knowledge or changing conditions [99,103]. Nevertheless, uncertainty is typically treated as a statistical performance measure rather than as an interoceptive signal integrated into broader regulatory dynamics governing internal stability and action selection.

In assistive and human-machine interaction systems, partial forms of interoceptive adaptation are already used in technologies that respond to physiological signals such as heart rate variability, fatigue indicators, or cognitive workload. Examples include adaptive rehabilitation systems, neurofeedback interfaces, and brain-computer interfaces that adjust task difficulty or feedback intensity according to the user's physiological state [128,129]. These approaches demonstrate the practical benefits of physiological adaptation in assistive technologies. However, interoception in such systems is largely externalized: physiological signals are used to adapt machine behavior without embedding analogous self-regulatory processes within the artificial agent itself.

Taken together, these examples illustrate that many ingredients of interoceptive regulation already exist across AI and robotics. The main limitation is not the absence of these mechanisms but their lack of integration within a coherent regulatory architecture. The interoceptive machine framework proposed here attempts to unify these components by treating internal state monitoring, anticipatory regulation, and action selection as interconnected processes supporting stability, adaptive behavior, and context-sensitive interaction.

Importantly, this proposal remains conceptual and architectural rather than fully validated empirically. Future work must therefore evaluate whether integrating these mechanisms into a unified interoceptive framework provides measurable advantages over existing modular approaches, particularly in long-horizon interactive settings such as assistive robotics, rehabilitation technologies, and human-machine interaction in general.

## 8. A path toward near-conscious AI?

As developed in previous sections, functionally grounded, context-sensitive behavior, the process by which an agent organizes behavior relative to internal and external constraints, requires more than just embodied interaction with the environment. Autonomous systems must generate their own identity and determine the significance of their interactions. They do not merely react to stimuli; they evaluate environmental changes in terms of how those changes affect their ongoing self-organization.

In this view, behavioral relevance emerges from the interaction between internal regulatory variables and environmental dynamics, shaped by what we can call the interoceptive-enactive principle. This principle integrates homeostatic and allostatic regulation. The relational domain that constitutes the agent's perspective is not pre-given but enacted through continuous activity. Relevance, therefore, is co-determined by the interplay between internal dynamics and external conditions.

The concept of autopoiesis [88], or the self-production of a living system, helps explain how identity is maintained over time. However, it is often framed as either a system maintains its organization, or it collapses [89]. This framing falls short in accounting for how an agent interprets diverse perturbations or flexibly adjusts its behavior to enhance survival. Adaptivity addresses this limitation by referring to a system's ability to monitor both internal and external states in relation to its viability, and to regulate behavior accordingly [89]. An adaptive system can assess whether ongoing trends are conducive to or threatening its continued existence and respond in ways that support its persistence, not through external commands but through internally driven evaluations.

Adaptivity, the capacity to regulate behavior based on internal assessments of viability, is central to both interoception and enactive cognition. Unlike externally dictated responses, adaptive behavior emerges from the system's own evaluations of how environmental changes affect its internal state—how much effort is required, what adjustments are needed, and whether these changes move the system toward or away from a viability threshold. This ability allows an agent to assign different valences to different situations, enriching its capacity for meaningful engagement. The interoceptive machine framework builds on this foundation [18] by offering a practical approach to enactive AI. It proposes systems that go beyond sensorimotor coupling, incorporating interoceptive processes in the self-constitution of autonomous identity. The framework emphasizes two key capacities: the role of interoception in constitutive autonomy, the generation and maintenance of identity through ongoing activity, and the role of adaptivity in responding flexibly to open-ended contexts without relying on pre-specified rules [78]. Together, these principles support the emergence of agents capable of internally regulated, context-dependent behavior, where meaning arises from internally regulated interactions rather than imposed behavior patterns.

Constitutive autonomy and adaptivity are essential, though not necessarily sufficient, conditions for intentional agency. While they lay the groundwork for systems that engage with the world in a purposeful and meaningful way, additional systemic properties may be needed to account for higher-level phenomena such as emotion, normativity, or social interaction [76,84,130]. Meaning would arise only when a system regulates its interactions in relation to its own identity, a process made possible by autonomy. Without this intrinsic perspective, it becomes difficult to determine whether a system is genuinely acting or merely being acted upon. Sense-making, therefore, requires both the ongoing production of identity and the dynamic modulation of environmental engagement[131]. This supports a central claim of enactive theory: sensorimotor interaction alone is not enough to ground intrinsic meaning and authentic agency [89].

Together, constitutive autonomy and adaptivity lay the conceptual foundation for the development of artificial agents that not only act in the world but do so from behavior governed by internally maintained and dynamically updated state variables.

Importantly, the interoceptive machine framework allows these internally regulated behaviors to be evaluated empirically, through metrics such as internal state estimation error, stability of viability variables, uncertainty calibration, and flexibility of internally driven goal adjustment.

## 9. Conclusion

Embodied AI, despite its advances, still falls short of capturing the essential qualities of human-like intentionality and meaning. A fundamental ontological gap persists between artificial systems and living organisms. Most notably, artificial systems lack a self-sustaining, self-producing mode of existence, often referred to as "being by doing." This limitation stems not only from technical challenges but also from the absence of a coherent theoretical foundation. However, recent developments in interoception research offer a promising basis for grounding such a foundation more firmly in biology and cognitive science.

In contrast, the biological view of cognition emphasizes its continuity with life. It places autonomous self-organization at the center of intelligent behavior. From this perspective, cognition is not simply about computation or representation, but about sense-making. Interoceptive-enactive AI builds on this view by proposing a transformative shift: the creation of systems that do not passively react to stimuli, but instead engage with their environments through internally regulated, meaning-driven interactions, with these internal (interoceptive) mechanisms playing a leading role.

Throughout this review, philosophical concepts were used to constrain and inform engineering design rather than to define engineering goals; the proposed interoceptive mechanisms are intended to be implementable and empirically evaluable, independent of unresolved questions about artificial consciousness. Nevertheless, this review also invites a deeper philosophical question: must consciousness be embodied? While the answer remains open, resolving this issue could have significant consequences for the development of AI and for determining which agents, biological or artificial, might possess conscious experience. The emergence of near-conscious AI would represent a major departure from classical models. It would move us toward systems that are not only adaptive and responsive, but also capable of self-regulation, self-assessment, and self-concern.

Importantly, the proposed framework does not start from a blank space but systematizes and extends regulatory mechanisms that already exist in contemporary AI, highlighting why their current fragmentation limits autonomy and how interoceptive integration could overcome these limitations. In this context, interoceptive AI frameworks represent a promising new frontier. By modeling the integration of internal and external signals—including multisensory inputs, affective states, and viability-based decision-making—these systems may offer valuable insights into the architecture of artificial agents or bionic systems. They point toward a future in which artificial agents are not merely functional tools, but participants in a meaningful world that they actively shape through their own perspective.

Importantly, the relevance of interoceptive regulation in assistive settings lies not in replacing human judgment, but in mitigating recurrent interaction failures by enabling artificial agents to recognize when to adapt, slow down, or disengage.

## 10. Acknowledgements

The author thanks Marie-Constance Corsi for the valuable feedback to an early version of this work.

## 11. References

[1] Park JS, O'Brien J, Cai CJ, Morris MR, Liang P, Bernstein MS. Generative Agents: Interactive Simulacra of Human Behavior. Proc. 36th Annu. ACM Symp. User Interface Softw. Technol., New York, NY, USA: Association for Computing Machinery; 2023, p. 1–22. https://doi.org/10.1145/3586183.3606763.
[2] Sap M, Le Bras R, Fried D, Choi Y. Neural Theory-of-Mind? On the Limits of Social Intelligence in Large LMs. In: Goldberg Y, Kozareva Z, Zhang Y, editors. Proc. 2022 Conf. Empir. Methods Nat. Lang. Process., Abu Dhabi, United Arab Emirates: Association for Computational Linguistics; 2022, p. 3762–80. https://doi.org/10.18653/v1/2022.emnlp-main.248.
[3] Aktan ME, Turhan Z, Dolu İ. Attitudes and perspectives towards the preferences for artificial intelligence in psychotherapy. Comput Hum Behav 2022;133:107273. https://doi.org/10.1016/j.chb.2022.107273.
[4] Chung EY-H, Kuen-Fung Sin K, Chow DH-K. Effectiveness of Robotic Intervention on Improving Social Development and Participation of Children with Autism Spectrum Disorder - A Randomised Controlled Trial. J Autism Dev Disord 2025;55:449–56. https://doi.org/10.1007/s10803-024-06236-2.
[5] Bharucha AJ, Anand V, Forlizzi J, Dew MA, Reynolds CF, Stevens S, et al. Intelligent Assistive Technology Applications to Dementia Care: Current Capabilities, Limitations, and Future Challenges. Am J Geriatr Psychiatry 2009;17:88–104. https://doi.org/10.1097/JGP.0b013e318187dde5.
[6] Varela FJ, Thompson E, Rosch E. The embodied mind: Cognitive science and human experience. Cambridge, MA, US: The MIT Press; 1991.
[7] Thompson E. Mind in Life: Biology, Phenomenology, and the Sciences of Mind. Harvard University Press; 2007.
[8] Rouleau N, Levin M. Discussions of machine versus living intelligence need more clarity. Nat Mach Intell 2024:1–3. https://doi.org/10.1038/s42256-024-00955-y.
[9] Pfeifer R, Lungarella M, Iida F. Self-Organization, Embodiment, and Biologically Inspired Robotics. Science 2007;318:1088–93. https://doi.org/10.1126/science.1145803.
[10] VanRullen R, Kanai R. Deep learning and the Global Workspace Theory. Trends Neurosci 2021;44:692–704. https://doi.org/10.1016/j.tins.2021.04.005.
[11] Storm JF, Klink PC, Aru J, Senn W, Goebel R, Pigorini A, et al. An integrative, multiscale view on neural theories of consciousness. Neuron 2024;112:1531–52. https://doi.org/10.1016/j.neuron.2024.02.004.
[12] Aru J, Larkum ME, Shine JM. The feasibility of artificial consciousness through the lens of neuroscience. Trends Neurosci 2023;46:1008–17. https://doi.org/10.1016/j.tins.2023.09.009.
[13] Gershman SJ. Uncertainty and exploration. Decision 2019;6:277–86. https://doi.org/10.1037/dec0000101.
[14] Pfeifer R, Gómez G. Interacting with the real world: design principles for intelligent systems. Artif Life Robot 2005;9:1–6. https://doi.org/10.1007/s10015-004-0343-3.
[15] Harvey I, Paolo ED, Wood R, Quinn M, Tuci E. Evolutionary Robotics: A New Scientific Tool for Studying Cognition. Artif Life 2005;11:79–98. https://doi.org/10.1162/1064546053278991.

[16]	Sharkey NE, Ziemke T. Mechanistic versus phenomenal embodiment: Can robot embodiment lead to strong AI? Cogn Syst Res 2001;2:251–62. https://doi.org/10.1016/S1389-0417(01)00036-5.
[17]	Anderson ML. Embodied Cognition: A field guide. Artif Intell 2003;149:91–130. https://doi.org/10.1016/S0004-3702(03)00054-7.
[18]	Froese T, Ziemke T. Enactive artificial intelligence: Investigating the systemic organization of life and mind. Artif Intell 2009;173:466–500. https://doi.org/10.1016/j.artint.2008.12.001.
[19]	Pfeifer R, Iida F, Bongard J. New Robotics: Design Principles for Intelligent Systems. Artif Life 2005;11:99–120. https://doi.org/10.1162/1064546053279017.
[20]	Pfeifer R, Bongard J. How the Body Shapes the Way We Think: A New View of Intelligence. n.d.
[21]	Dreyfus HL. Why Heideggerian AI Failed and How Fixing it Would Require Making it More Heideggerian. Philos Psychol 2007;20:247–68. https://doi.org/10.1080/09515080701239510.
[22]	Dennett DC. Cognitive wheels: The frame problem of AI. New York, NY, US: Routledge/Taylor & Francis Group; 2006.
[23]	Di Paolo EA, Iizuka H. How (not) to model autonomous behaviour. Biosystems 2008;91:409–23. https://doi.org/10.1016/j.biosystems.2007.05.016.
[24]	Block N. On a confusion about a function of consciousness. Behav Brain Sci 1995;18:227–47. https://doi.org/10.1017/S0140525X00038188.
[25]	Shea N, Boldt A, Bang D, Yeung N, Heyes C, Frith CD. Supra-personal cognitive control and metacognition. Trends Cogn Sci 2014;18:186–93. https://doi.org/10.1016/j.tics.2014.01.006.
[26]	Barrett LF, Simmons WK. Interoceptive predictions in the brain. Nat Rev Neurosci 2015;16:419–29. https://doi.org/10.1038/nrn3950.
[27]	Chen WG, Schloesser D, Arensdorf AM, Simmons JM, Cui C, Valentino R, et al. The Emerging Science of Interoception: Sensing, Integrating, Interpreting, and Regulating Signals within the Self. Trends Neurosci 2021;44:3–16. https://doi.org/10.1016/j.tins.2020.10.007.
[28]	Candia-Rivera D. Brain-heart interactions in the neurobiology of consciousness. Curr Res Neurobiol 2022;3:100050. https://doi.org/10.1016/j.crneur.2022.100050.
[29]	Man K, Damasio A. Homeostasis and soft robotics in the design of feeling machines. Nat Mach Intell 2019;1:446–52. https://doi.org/10.1038/s42256-019-0103-7.
[30]	Taniguchi T, Murata S, Suzuki M, Ognibene D, Lanillos P, Ugur E, et al. World models and predictive coding for cognitive and developmental robotics: frontiers and challenges. Adv Robot 2023;37:780–806. https://doi.org/10.1080/01691864.2023.2225232.
[31]	Juechems K, Balaguer J, Spitzer B, Summerfield C. Optimal utility and probability functions for agents with finite computational precision. Proc Natl Acad Sci 2021;118:e2002232118. https://doi.org/10.1073/pnas.2002232118.
[32]	Keramati M, Gutkin B. Homeostatic reinforcement learning for integrating reward collection and physiological stability. eLife 2014;3:e04811. https://doi.org/10.7554/eLife.04811.
[33]	Hu Q, Ji L, Wang Y, Zhao S, Lin Z. Uncertainty-driven active developmental learning. Pattern Recognit 2024;151:110384. https://doi.org/10.1016/j.patcog.2024.110384.
[34]	Singh S, Lewis RL, Barto AG, Sorg J. Intrinsically Motivated Reinforcement Learning: An Evolutionary Perspective. IEEE Trans Auton Ment Dev 2010;2:70–82. https://doi.org/10.1109/TAMD.2010.2051031.
[35]	Hulme OJ, Morville T, Gutkin B. Neurocomputational theories of homeostatic control. Phys Life Rev 2019;31:214–32. https://doi.org/10.1016/j.plrev.2019.07.005.
[36]	Fouragnan EF, Hosking B, Cheung Y, Prakash B, Rushworth M, Sel A. Timing along the cardiac cycle modulates neural signals of reward-based learning. Nat Commun 2024;15:2976. https://doi.org/10.1038/s41467-024-46921-5.
[37]	Cardenas MA, Le RP, Champ TM, O'Neill D, Fuglevand AJ, Gothard KM. Manipulation of interoceptive signaling biases decision making in rhesus macaques. Proc Natl Acad Sci 2025;122:e2424680122. https://doi.org/10.1073/pnas.2424680122.
[38]	Palser ER, Glass J, Fotopoulou A, Kilner JM. Relationship between cardiac cycle and the timing of actions during action execution and observation. Cognition 2021;217:104907. https://doi.org/10.1016/j.cognition.2021.104907.
[39]	Azzalini D, Buot A, Palminteri S, Tallon-Baudry C. Responses to heartbeats in ventromedial prefrontal cortex contribute to subjective preference-based decisions. J Neurosci 2021. https://doi.org/10.1523/JNEUROSCI.1932-20.2021.
[40]	Fujimoto A, Murray EA, Rudebeck PH. Interaction between decision-making and interoceptive representations of bodily arousal in frontal cortex. Proc Natl Acad Sci 2021;118. https://doi.org/10.1073/pnas.2014781118.
[41]	Soori M, Arezoo B, Dastres R. Optimization of energy consumption in industrial robots, a review. Cogn Robot 2023;3:142–57. https://doi.org/10.1016/j.cogr.2023.05.003.
[42]	Seth AK, Tsakiris M. Being a Beast Machine: The Somatic Basis of Selfhood. Trends Cogn Sci

2018;22:969–81. https://doi.org/10.1016/j.tics.2018.08.008.
[43] Lee S, Oh Y, An H, Yoon H, Friston KJ, Hong SJ, et al. Life-inspired Interoceptive Artificial Intelligence for Autonomous and Adaptive Agents 2023. https://doi.org/10.48550/arXiv.2309.05999.
[44] Mohr PNC, Biele G, Heekeren HR. Neural Processing of Risk. J Neurosci 2010;30:6613–9. https://doi.org/10.1523/JNEUROSCI.0003-10.2010.
[45] Hernández A, Zainos A, Romo R. Temporal Evolution of a Decision-Making Process in Medial Premotor Cortex. Neuron 2002;33:959–72. https://doi.org/10.1016/S0896-6273(02)00613-X.
[46] Kashi Y, King DG. Simple sequence repeats as advantageous mutators in evolution. Trends Genet 2006;22:253–9. https://doi.org/10.1016/j.tig.2006.03.005.
[47] Hylin MJ, Kerr AL, Holden R. Understanding the Mechanisms of Recovery and/or Compensation following Injury. Neural Plast 2017;2017:7125057. https://doi.org/10.1155/2017/7125057.
[48] Sterling P, Laughlin S. Principles of Neural Design. The MIT Press; 2015. https://doi.org/10.7551/mitpress/9780262028707.001.0001.
[49] Su H, Ovur SE, Xu Z, Alfayad S. Exploring the Potential of Fuzzy Sets in Cyborg Enhancement: A Comprehensive Review. IEEE Trans Fuzzy Syst 2025;33:810–27. https://doi.org/10.1109/TFUZZ.2024.3491733.
[50] Petzschner FH, Garfinkel SN, Paulus MP, Koch C, Khalsa SS. Computational Models of Interoception and Body Regulation. Trends Neurosci 2021;44:63–76. https://doi.org/10.1016/j.tins.2020.09.012.
[51] Gershman SJ. Deconstructing the human algorithms for exploration. Cognition 2018;173:34–42. https://doi.org/10.1016/j.cognition.2017.12.014.
[52] Aston-Jones G, Cohen JD. An integrative theory of locus coeruleus-norepinephrine function: adaptive gain and optimal performance. Annu Rev Neurosci 2005;28:403–50. https://doi.org/10.1146/annurev.neuro.28.061604.135709.
[53] Rudebeck PH, Izquierdo A. Foraging with the frontal cortex: A cross-species evaluation of reward-guided behavior. Neuropsychopharmacology 2022;47:134–46. https://doi.org/10.1038/s41386-021-01140-0.
[54] Haselager WFG. Robotics, philosophy and the problems of autonomy. Pragmat Cogn 2005;13:515–32. https://doi.org/10.1075/pc.13.3.07has.
[55] Moreno A, Etxeberria A. Agency in Natural and Artificial Systems. Artif Life 2005;11:161–75. https://doi.org/10.1162/1064546053278919.
[56] Harnad S. The symbol grounding problem. Phys Nonlinear Phenom 1990;42:335–46. https://doi.org/10.1016/0167-2789(90)90087-6.
[57] Adaptive Agent Team, Bauer J, Baumli K, Baveja S, Behbahani F, Bhoopchand A, et al. Human-Timescale Adaptation in an Open-Ended Task Space 2023. https://doi.org/10.48550/arXiv.2301.07608.
[58] Tallon-Baudry C. Interoception: Probing internal state is inherent to perception and cognition. Neuron 2023;111:1854–7. https://doi.org/10.1016/j.neuron.2023.04.019.
[59] Hsueh B, Chen R, Jo Y, Tang D, Raffiee M, Kim YS, et al. Cardiogenic control of affective behavioural state. Nature 2023;615:292–9. https://doi.org/10.1038/s41586-023-05748-8.
[60] Hill MW, Johnson E, Ellmers TJ. The influence of false interoceptive feedback on emotional state and balance responses to height-induced postural threat. Biol Psychol 2024;189:108803. https://doi.org/10.1016/j.biopsycho.2024.108803.
[61] Candia-Rivera D, Catrambone V, Thayer JF, Gentili C, Valenza G. Cardiac sympathetic-vagal activity initiates a functional brain–body response to emotional arousal. Proc Natl Acad Sci 2022;119:e2119599119. https://doi.org/10.1073/pnas.2119599119.
[62] Payzan-LeNestour E, Dunne S, Bossaerts P, O'Doherty JP. The Neural Representation of Unexpected Uncertainty during Value-Based Decision Making. Neuron 2013;79:191–201. https://doi.org/10.1016/j.neuron.2013.04.037.
[63] Allen M, Frank D, Schwarzkopf DS, Fardo F, Winston JS, Hauser TU, et al. Unexpected arousal modulates the influence of sensory noise on confidence. eLife 2016;5:e18103. https://doi.org/10.7554/eLife.18103.
[64] Sennesh E, Theriault J, Brooks D, van de Meent J-W, Barrett LF, Quigley KS. Interoception as modeling, allostasis as control. Biol Psychol 2022;167:108242. https://doi.org/10.1016/j.biopsycho.2021.108242.
[65] Kleckner IR, Zhang J, Touroutoglou A, Chanes L, Xia C, Simmons WK, et al. Evidence for a Large-Scale Brain System Supporting Allostasis and Interoception in Humans. Nat Hum Behav 2017;1:0069. https://doi.org/10.1038/s41562-017-0069.
[66] Matsuo Y, LeCun Y, Sahani M, Precup D, Silver D, Sugiyama M, et al. Deep learning, reinforcement learning, and world models. Neural Netw 2022;152:267–75. https://doi.org/10.1016/j.neunet.2022.03.037.
[67] Weyns D, Calinescu R, Mirandola R, Tei K, Acosta M, Bencomo N, et al. Towards a Research Agenda for Understanding and Managing Uncertainty in Self-Adaptive Systems. SIGSOFT Softw Eng Notes 2023;48:20–36. https://doi.org/10.1145/3617946.3617951.
[68] Schulkin J, Sterling P. Allostasis: A Brain-Centered, Predictive Mode of Physiological Regulation. Trends Neurosci 2019;42:740–52. https://doi.org/10.1016/j.tins.2019.07.010.

[69]     Salvatori T, Mali A, Buckley CL, Lukasiewicz T, Rao RPN, Friston K, et al. A survey on neuro-mimetic deep learning via predictive coding. Neural Netw 2026;195:108161. https://doi.org/10.1016/j.neunet.2025.108161.
[70]     Kendall A, Gal Y. What Uncertainties Do We Need in Bayesian Deep Learning for Computer Vision? Adv. Neural Inf. Process. Syst., vol. 30, Curran Associates, Inc.; 2017.
[71]     Wilson RC, Collins AG. Ten simple rules for the computational modeling of behavioral data. eLife 2019;8:e49547. https://doi.org/10.7554/eLife.49547.
[72]     Dayan P. Motivated Reinforcement Learning. Adv. Neural Inf. Process. Syst., vol. 14, MIT Press; 2001.
[73]     Corrales-Carvajal VM, Faisal AA, Ribeiro C. Internal states drive nutrient homeostasis by modulating exploration-exploitation trade-off. eLife 2016;5:e19920. https://doi.org/10.7554/eLife.19920.
[74]     Flavell S, Gogolla N, Lovett-Barron M, Zelikowsky M. The Emergence and Influence of Internal States. Neuron 2022;110:2545–70. https://doi.org/10.1016/j.neuron.2022.04.030.
[75]     Weber A, Varela FJ. Life after Kant: Natural purposes and the autopoietic foundations of biological individuality. Phenomenol Cogn Sci 2002;1:97–125. https://doi.org/10.1023/A:1020368120174.
[76]     Ziemke T. On the role of emotion in biological and robotic autonomy. Biosystems 2008;91:401–8. https://doi.org/10.1016/j.biosystems.2007.05.015.
[77]     O'Regan JK, Noë A. A sensorimotor account of vision and visual consciousness. Behav Brain Sci 2001;24:939–73. https://doi.org/10.1017/S0140525X01000115.
[78]     Jonas H. The Phenomenon of Life: Toward a Philosophical Biology. Northwestern University Press; 2001.
[79]     Thompson E. Sensorimotor subjectivity and the enactive approach to experience. Phenomenol Cogn Sci 2005;4:407–27. https://doi.org/10.1007/s11097-005-9003-x.
[80]     Kudithipudi D, Aguilar-Simon M, Babb J, Bazhenov M, Blackiston D, Bongard J, et al. Biological underpinnings for lifelong learning machines. Nat Mach Intell 2022;4:196–210. https://doi.org/10.1038/s42256-022-00452-0.
[81]     Candia-Rivera D, Engelen T, Babo-Rebelo M, Salamone PC. Interoception, Network Physiology and the Emergence of Bodily Self-Awareness. Neurosci Biobehav Rev 2024;165:105864. https://doi.org/10.1016/j.neubiorev.2024.105864.
[82]     De Loor P, Manac'h K, Tisseau J. Enaction-Based Artificial Intelligence: Toward Co-evolution with Humans in the Loop. Minds Mach 2009;19:319–43. https://doi.org/10.1007/s11023-009-9165-3.
[83]     Friston KJ, Rosch R, Parr T, Price C, Bowman H. Deep temporal models and active inference. Neurosci Biobehav Rev 2017;77:388–402. https://doi.org/10.1016/j.neubiorev.2017.04.009.
[84]     Duéñez-Guzmán EA, Sadedin S, Wang JX, McKee KR, Leibo JZ. A social path to human-like artificial intelligence. Nat Mach Intell 2023;5:1181–8. https://doi.org/10.1038/s42256-023-00754-x.
[85]     Bolis D, Schilbach L. 'I Interact Therefore I Am': The Self as a Historical Product of Dialectical Attunement. Topoi 2020;39:521–34. https://doi.org/10.1007/s11245-018-9574-0.
[86]     Houthooft R, Chen X, Chen X, Duan Y, Schulman J, De Turck F, et al. VIME: Variational Information Maximizing Exploration. Adv. Neural Inf. Process. Syst., vol. 29, Curran Associates, Inc.; 2016.
[87]     Tang W, Zhu M, Wu F, Li X, Liu Y. Empowering Embodied Agents with Semantic Intelligence. Eng. Complex Comput. Syst. 29th Int. Conf. ICECCS 2025 Hangzhou China July 2–4 2025 Proc., Berlin, Heidelberg: Springer-Verlag; 2025, p. 480–5. https://doi.org/10.1007/978-3-032-00828-2_27.
[88]     Varela FJ, Maturana HR, Uribe R. Autopoiesis: The organization of living systems, its characterization and a model. Biosystems 1974;5:187–96. https://doi.org/10.1016/0303-2647(74)90031-8.
[89]     Di Paolo EA. Autopoiesis, Adaptivity, Teleology, Agency. Phenomenol Cogn Sci 2005;4:429–52. https://doi.org/10.1007/s11097-005-9002-y.
[90]     Beer JM, Fisk AD, Rogers WA. Toward a framework for levels of robot autonomy in human-robot interaction. J Hum-Robot Interact 2014;3:74–99. https://doi.org/10.5898/JHRI.3.2.Beer.
[91]     Oudeyer P-Y, Kaplan F. What is Intrinsic Motivation? A Typology of Computational Approaches. Front Neurorobotics 2007;1:6. https://doi.org/10.3389/neuro.12.006.2007.
[92]     Howard RA, Matheson JE. Risk-Sensitive Markov Decision Processes. Manag Sci 1972;18:356–69.
[93]     Bellemare MG, Dabney W, Munos R. A Distributional Perspective on Reinforcement Learning. Proc. 34th Int. Conf. Mach. Learn. - Vol. 70, Sydney, NSW, Australia: JMLR.org; 2017, p. 449–58.
[94]     Osband I, Blundell C, Pritzel A, Van Roy B. Deep Exploration via Bootstrapped DQN. Adv. Neural Inf. Process. Syst., vol. 29, Curran Associates, Inc.; 2016.
[95]     Sterling P. Allostasis: A model of predictive regulation. Physiol Behav 2012;106:5–15. https://doi.org/10.1016/j.physbeh.2011.06.004.
[96]     Pathak D, Agrawal P, Efros AA, Darrell T. Curiosity-driven exploration by self-supervised prediction. Proc. 34th Int. Conf. Mach. Learn. - Vol. 70, Sydney, NSW, Australia: JMLR.org; 2017, p. 2778–87.
[97]     Salge C, Glackin C, Polani D. Empowerment–An Introduction. In: Prokopenko M, editor. Guid. Self-Organ. Inception, Berlin, Heidelberg: Springer; 2014, p. 67–114. https://doi.org/10.1007/978-3-642-53734-9_4.

[98] Guo C, Pleiss G, Sun Y, Weinberger KQ. On Calibration of Modern Neural Networks. Proc. 34th Int. Conf. Mach. Learn., PMLR; 2017, p. 1321–30.
[99] Lakshminarayanan B, Pritzel A, Blundell C. Simple and Scalable Predictive Uncertainty Estimation using Deep Ensembles. Adv. Neural Inf. Process. Syst., vol. 30, Curran Associates, Inc.; 2017.
[100] Geifman Y, El-Yaniv R. Selective classification for deep neural networks. Proc. 31st Int. Conf. Neural Inf. Process. Syst., Red Hook, NY, USA: Curran Associates Inc.; 2017, p. 4885–94.
[101] Hendrycks D, Basart S, Mazeika M, Zou A, Kwon J, Mostajabi M, et al. Scaling Out-of-Distribution Detection for Real-World Settings. Proc. 39th Int. Conf. Mach. Learn., PMLR; 2022, p. 8759–73.
[102] Liu W, Wang X, Owens J, Li Y. Energy-based Out-of-distribution Detection. Adv. Neural Inf. Process. Syst., vol. 33, Curran Associates, Inc.; 2020, p. 21464–75.
[103] Ovadia Y, Fertig E, Ren J, Nado Z, Sculley D, Nowozin S, et al. Can you trust your model' s uncertainty? Evaluating predictive uncertainty under dataset shift. Adv. Neural Inf. Process. Syst., vol. 32, Curran Associates, Inc.; 2019.
[104] Krueger D, Caballero E, Jacobsen J-H, Zhang A, Binas J, Zhang D, et al. Out-of-Distribution Generalization via Risk Extrapolation (REx). Proc. 38th Int. Conf. Mach. Learn., PMLR; 2021, p. 5815–26.
[105] Scherer KR. Appraisal considered as a process of multilevel sequential checking. Apprais. Process. Emot. Theory Methods Res., New York, NY, US: Oxford University Press; 2001, p. 92–120.
[106] Moors A, Ellsworth PC, Scherer KR, Frijda NH. Appraisal theories of emotion: State of the art and future development. Emot Rev 2013;5:119–24. https://doi.org/10.1177/1754073912468165.
[107] Marsella SC, Gratch J. EMA: A process model of appraisal dynamics. Cogn Syst Res 2009;10:70–90. https://doi.org/10.1016/j.cogsys.2008.03.005.
[108] Picard RW. Affective computing. Cambridge, MA, US: The MIT Press; 1997.
[109] Yu AJ, Dayan P. Uncertainty, neuromodulation, and attention. Neuron 2005;46:681–92. https://doi.org/10.1016/j.neuron.2005.04.026.
[110] Friston K. The free-energy principle: a unified brain theory? Nat Rev Neurosci 2010;11:127–38. https://doi.org/10.1038/nrn2787.
[111] Seth AK. Interoceptive inference, emotion, and the embodied self. Trends Cogn Sci 2013;17:565–73. https://doi.org/10.1016/j.tics.2013.09.007.
[112] García J, Fernández F. A Comprehensive Survey on Safe Reinforcement Learning. J Mach Learn Res 2015;16:1437–80.
[113] Dulac-Arnold G, Levine N, Mankowitz DJ, Li J, Paduraru C, Gowal S, et al. Challenges of real-world reinforcement learning: definitions, benchmarks and analysis. Mach Learn 2021;110:2419–68. https://doi.org/10.1007/s10994-021-05961-4.
[114] Amodei D, Olah C, Steinhardt J, Christiano P, Schulman J, Mané D. Concrete Problems in AI Safety 2016. https://doi.org/10.48550/arXiv.1606.06565.
[115] Kompella VR, Stollenga M, Luciw M, Schmidhuber J. Continual curiosity-driven skill acquisition from high-dimensional video inputs for humanoid robots. Artif Intell 2017;247:313–35. https://doi.org/10.1016/j.artint.2015.02.001.
[116] Klyubin AS, Polani D, Nehaniv CL. Empowerment: a universal agent-centric measure of control. 2005 IEEE Congr. Evol. Comput., vol. 1, 2005, p. 128-135 Vol.1. https://doi.org/10.1109/CEC.2005.1554676.
[117] Schmidhuber J. Formal Theory of Creativity, Fun, and Intrinsic Motivation (1990–2010). IEEE Trans Auton Ment Dev 2010;2:230–47. https://doi.org/10.1109/TAMD.2010.2056368.
[118] Waytz A, Heafner J, Epley N. The mind in the machine: Anthropomorphism increases trust in an autonomous vehicle. J Exp Soc Psychol 2014;52:113–7. https://doi.org/10.1016/j.jesp.2014.01.005.
[119] Kim Y, Sundar SS. Anthropomorphism of computers: Is it mindful or mindless? Comput Hum Behav 2012;28:241–50. https://doi.org/10.1016/j.chb.2011.09.006.
[120] Floridi L. Establishing the Rules for Building Trustworthy AI. In: Floridi L, editor. Ethics Gov. Policies Artif. Intell., Cham: Springer International Publishing; 2021, p. 41–5. https://doi.org/10.1007/978-3-030-81907-1_4.
[121] Fitzpatrick KK, Darcy A, Vierhile M. Delivering Cognitive Behavior Therapy to Young Adults With Symptoms of Depression and Anxiety Using a Fully Automated Conversational Agent (Woebot): A Randomized Controlled Trial. JMIR Ment Health 2017;4:e7785. https://doi.org/10.2196/mental.7785.
[122] Scassellati B, Admoni H, Matarić M. Robots for use in autism research. Annu Rev Biomed Eng 2012;14:275–94. https://doi.org/10.1146/annurev-bioeng-071811-150036.
[123] Pennisi P, Tonacci A, Tartarisco G, Billeci L, Ruta L, Gangemi S, et al. Autism and social robotics: A systematic review. Autism Res Off J Int Soc Autism Res 2016;9:165–83. https://doi.org/10.1002/aur.1527.
[124] Topo P. Technology Studies to Meet the Needs of People With Dementia and Their Caregivers: A Literature Review. J Appl Gerontol 2009;28:5–37. https://doi.org/10.1177/0733464808324019.
[125] Bemelmans R, Gelderblom GJ, Jonker P, de Witte L. Socially assistive robots in elderly care: a systematic review into effects and effectiveness. J Am Med Dir Assoc 2012;13:114-120.e1.

https://doi.org/10.1016/j.jamda.2010.10.002.

[126] Krakovna V, Orseau L, Ngo R, Martic M, Legg S. Avoiding Side Effects By Considering Future Tasks. Adv. Neural Inf. Process. Syst., vol. 33, Curran Associates, Inc.; 2020, p. 19064–74.

[127] Kober J, Peters J. Reinforcement Learning in Robotics: A Survey. In: Kober J, Peters J, editors. Learn. Mot. Ski. Algorithms Robot Exp., Cham: Springer International Publishing; 2014, p. 9–67. https://doi.org/10.1007/978-3-319-03194-1_2.

[128] Novak D, Ziherl J, Olensek A, Milavec M, Podobnik J, Mihelj M, et al. Psychophysiological responses to robotic rehabilitation tasks in stroke. IEEE Trans Neural Syst Rehabil Eng Publ IEEE Eng Med Biol Soc 2010;18:351–61. https://doi.org/10.1109/TNSRE.2010.2047656.

[129] Fairclough SH. Fundamentals of physiological computing. Interact Comput 2009;21:133–45. https://doi.org/10.1016/j.intcom.2008.10.011.

[130] De Jaegher H, Di Paolo E. Participatory sense-making. Phenomenol Cogn Sci 2007;6:485–507. https://doi.org/10.1007/s11097-007-9076-9.

[131] Bourgine P, Stewart J. Autopoiesis and Cognition. Artif Life 2004;10:327–45. https://doi.org/10.1162/1064546041255557.